\documentclass{article}

\usepackage[accepted]{aistats2020}

\usepackage[utf8]{inputenc} 
\usepackage[T1]{fontenc}    
\usepackage{hyperref}       
\usepackage{url}            
\usepackage{booktabs}       
\usepackage{amsfonts}       
\usepackage{nicefrac}       
\usepackage{microtype}      

\usepackage[pdftex]{graphicx}
\usepackage{wrapfig}
\usepackage{amsmath,amsthm}
\usepackage{subcaption}

\usepackage[style=authoryear,uniquename=false,natbib]{biblatex} 

\usepackage{xcolor}

\newtheorem{theorem}{Theorem}[section]

\newtheorem{definition}[theorem]{Definition}

\newcommand{\PD}{\mathrm{PD}}
\newcommand{\RR}{\mathbb{R}}
\newcommand{\calI}{\mathcal{I}}
\newcommand{\calE}{\mathcal{E}}
\newcommand{\II}{\mathbb{I}}

\newcommand{\X}{\mathcal{X}}
\DeclareMathOperator{\im}{im}

\title{A Topology Layer for Machine Learning}

\begin{document}

\twocolumn[

\aistatstitle{A Topology Layer for Machine Learning}


 \aistatsauthor{ Rickard Brüel-Gabrielsson \And Bradley J. Nelson \And Anjan Dwaraknath}
 \aistatsaddress{ Stanford University, UnboxAI \And Stanford University \And Stanford University}

 \aistatsauthor{ Primoz Skraba \And Leonidas J. Guibas \And Gunnar Carlsson}

 \aistatsaddress{ Queen Mary University of London  \And  Stanford University  \And Stanford University, UnboxAI } ]

\begin{abstract}
Topology applied to real world data using persistent homology has started to find applications within machine learning, including deep learning. We present a differentiable topology layer that computes persistent homology based on level set filtrations and edge-based filtrations. We present three  novel applications: the topological layer can (i) regularize data reconstruction or the weights of machine learning models, (ii) construct a loss on the output of a deep generative network to incorporate topological priors, and (iii) perform topological adversarial attacks on deep networks trained with persistence features. The code\footnote{\url{www.github.com/bruel-gabrielsson/TopologyLayer}} is publicly available and we hope its availability will facilitate the use of persistent homology in deep learning and other gradient based applications. 
\end{abstract}

\section{Introduction}
\label{sec:intro}
Persistent homology, or simply persistence, is a well-established tool in applied and computational topology. In a deep learning setting, persistence has mainly been used as preprocessing to provide topological features for learning~\citep{persistsurvey, persist1, persist2}. There has been work that uses differentiable properties of persistence to incorporate topological information in deep learning~\citep{priors, Music} and regularization~\citep{TopReg}; however, such work has focused on specialized applications and specific functions of the persistence diagrams. Another line of work uses applied topology and persistence for deep learning interpretability~\citep{TopExp, TopLearn}, automating the construction of deep learning architectures~\citep{TopDeep}, complexity measures~\citep{complex1, complex2}, and adversarial attacks~\citep{geb1, geb2}. Persistence fits naturally in geometric problems and has been applied to a number of geometry processing applications including shape matching~\citep{carlsson2005persistence}, optimal pose-matching~\citep{dey2010persistent}, shape segmentation~\citep{skraba2010persistence}, and surface reconstruction~\citep{rbgsurface}. In~\citep{rbgsurface} gradient descent was successfully applied to persistence-based optimization.  In this work, we consider how gradient descent through persistence may be used more broadly and flexibly than in the specialized applications cited above.  As we will see, persistent homology is easily tailored to incorporate topological information into a variety of machine learning problems.

In many deep learning settings there is a natural topological perspective, including images and for 3D data such as point clouds or voxel spaces. In fact, many of the failure cases of generative models are topological in nature~\citep{3dgan, GAN}. We show how topological priors can be used to improve such models. It has been speculated~\citep{TopExp, TopDeep} that models that rely on topological features might have desirable properties besides test accuracy; one such property has been robustness against adversarial attacks~\citep{advSurvey}. However, to our knowledge, no such attacks have been conducted. With our layer, such attacks are easy to implement, and we provide illustrative examples. As a language for describing global properties of data, topology is also useful in exploring properties of generalization. There are some natural measures of topological simplicity (akin to norm regularization) and we show how these can successfully be used to regularize the parameters or weights of machine learning models.
Our contributions include: (i) an easy-to-use persistence layer for level set filtrations and edge-based filtrations, (ii) the first regularization using persistence directly on the weights of machine learning models, (iii) the first incorporation of topological priors in deep generative networks in image and 3D data settings, and (iv) the first topological adversarial attacks.

\section{Topological Preliminaries}
\label{sec:persistence}

This section contains a review of the relevant topological notions, including persistent homology (persistence), providing an intuitive idea with some additional details provided in the supplementary material. For readers  who are unfamiliar with persistence,  we refer the reader to a number of excellent introductions and surveys which are available~\citep{books/daglib/0025666} and~\citep{poulenard2018topological,rbgsurface} for related work on optimizing over persistence diagrams.

Topological spaces can generally be encoded using cell complexes, which consist of $k$-dimensional balls, or cells, ($k=0,1,2,\dots$), and boundary maps from cells in dimension $k$
to cells in dimension $k-1$.  Practically, it is convenient to work with simplicial complexes, in which $k$-dimensional cells are $k$-dimensional simplices, because boundary maps are determined automatically, but we will continue this section by discussing more general cell complexes.  We will assume that the complexes are finite which ensures various technical conditions necessary for persistence \citep{books/daglib/0025666}.

{\bf Homology} is an {\em algebraic invariant} of a topological space, associating a vector space $H_k$ to the $k$th dimension of a cell complex $\X$.  Homology is computed by forming a chain complex of $\X$, consisting of vector spaces $C_k(\X)$ which are freely generated by the $k$-cells of $\X$, and boundary maps $\partial_k:C_k(\X) \to C_{k-1}(\X)$ which satisfy $\partial_{k-1} \circ \partial_k = 0$.  In the case of dimension 0, $\partial_0 = 0$.  As a notational convenience, we will use the same symbol for a $k$-cell $\sigma\in \X$ and the associated basis vector $\sigma\in C_k(\X)$. The vector spaces $C_k(\X)$ may be over any field, but for the purposes of determining kernels and images of maps exactly finite fields are preferred -- in practice we use the finite field with two elements, $\mathbb{Z}/2\mathbb{Z}$. Homology in dimension $k$ is defined as the quotient vector space
$$H_k(\X) = \ker \partial_k / \im \partial_{k+1}$$
An element of $H_k(\X)$ is called a homology (equivalence) class, and a choice of representative for a class is called a generator.  The dimension of $H_k(\X)$ counts the number of $k$-dimensional features of $\X$.  For example, $\dim H_0(\X)$ counts the number of connected components, $\dim H_1(\X)$ counts the number of holes, and so on.  Homology is {\em homotopy invariant}, meaning that continuous deformations of $\X$ produce the same result.

{\bf Persistent homology} studies how homology changes over an increasing sequence of complexes $\X_0 \subseteq \X_1 \subseteq \dots \subseteq \X_n = \X$, also called a filtration on $\X$.  We consider sublevel set filtrations of a function $f:\X \rightarrow \mathbb{R}$. The filtration is defined by increasing the parameter $\alpha$, with $\X_\alpha = f^{-1}(-\infty,\alpha]$,
and the only requirement is that $\X_\alpha$ be a valid cell complex, meaning if a cell is in $\X_\alpha$, its boundary must also be in $\X_\alpha$.  We first consider a filtration where cells have a strict ordering, meaning they are added one at a time.  The addition of a $k$-dimensional cell $\sigma$ at parameter $i$ can have two outcomes. First, if $\partial_k \sigma$ is already in $\im \partial_k$ (meaning $\partial_k \sigma = \partial_k w$ for some $w\in C_k(\X_{i} \setminus \sigma)$), then $w - \sigma \in \ker \partial_k$.  Since the kernel expands by one dimension, the quotient $H_k = \ker \partial_k / \im \partial_{k+1}$ expands by one dimension, and $w - \sigma$ generates the new homology class.  The second possibility is that $\partial_k \sigma$ is not already in $\im \partial_k$, which means $\im \partial_k$ expands by one dimension.  Because $\partial_{k-1} \circ \partial_k = 0$, $\partial_k \sigma \in \ker \partial_{k-1}$, and previously generated a homology class.  Thus, the quotient $H_{k-1} = \ker \partial_{k-1} / \im \partial_k$ will have one fewer dimension, and $\partial_k \sigma$ is a generator for the removed class.  In summary, every cell in the filtration either {\em creates} or {\em destroys} homology when it appears.
The full information about how homology is born and dies over the filtration can be represented as a multi-set of pairs $(b,d)$ where $b$ is the birth parameter of a homology class, and $d$ is the death parameter of that class ($d = \infty$ if it is still present in $\X$).  This multiset of pairs for homology in dimension $k$ is known as the $k$-dimensional persistence diagram of the filtration, $\PD_k(\X_\alpha) = \{(b_i, d_i)\}_{i\in \calI_k}$, or the $k$-dimensional barcode of $\X_\alpha$.  As a notational convenience, we will order the indexing of points by decreasing lifetimes i.e. $d_i - b_i \ge d_j - b_j$ for $i < j$.  

As persistence diagrams are a collection of points in $\mathbb{R}^2$, there are many notions of distances between diagrams and cost functions on diagrams which depend on the points. 
We use loss functions that can be expressed in terms of three parameters
\begin{equation}
\label{eq:cost}
\mathcal{E}(p, q, i_0; \PD_k) = \sum_{i= i_0}^{|\calI_k|} |d_i - b_i|^p (\tfrac{d_i + b_i}{2})^q
\end{equation}
The parameters $p$ and $q$ define a polynomial function, following those introduced in \citep{adcockRingAlgebraicFunctions2016}. We sum over lifetimes beginning with the $i_0$ most persistent point in the diagram. 
Varying $i_0$ for $\PD_k$ varies the number of $k$-dimensional features that are not penalized. For example, if $i_0 = 2$, with $\PD_0$ we consider all but the most persistent class, promoting a one connected component. Alternatively, using $i_0=2$ with $\PD_1$ will promote one hole.
The parameter $p$ can be increased to more strongly penalize the most persistent features, and the parameter $q$ serves to weight features that are prominent later in the filtration.
We also use the Wasserstein distance between diagrams -- this is defined as the optimal transport distance between the points of the two diagrams. One technicality is that the two diagrams may have different cardinalities, and points may be mapped to the diagonal -- see the supplementary material for details.

{\bf Differentiation:} Given an input filtration $f:\X \to \RR$, we can compute the gradient of a functional of a persistence diagram $\calE(\PD_k)$.  The key is to note that each birth-death pair can be mapped to the cells that respectively created and destroyed the homology class, defining an inverse map
\begin{equation}\label{eq:inversemap}
\pi_f(k) :   \{b_i,d_i\}_{i\in\mathcal{I}_k} \rightarrow (\sigma,\tau).
\end{equation}
In the case where the ordering on cells is strict, as we previously discussed, the map is unique, and we obtain
\begin{equation}\label{eq:derivative}
\frac{\partial \calE}{\partial \sigma} = \sum_{i\in I_k} \frac{\partial \calE}{\partial b_i}\II_{\pi_f(k)(b_i) = \sigma} + \sum_{i\in I_k} \frac{\partial \calE}{\partial d_i}\II_{\pi_f(k)(d_i) = \sigma}
\end{equation}
in which at most one term will have a non-zero indicator.  As we will see, many filtrations do not give rise to a strict ordering, because multiple cells can appear at the same parameter value in the filtration.  While the persistence diagram is still well-defined, the inverse map \ref{eq:inversemap} may no longer be unique.  This can be resolved by extending the total order to a strict order either deterministically or randomly -- see \citep{skraba2017randomly} for a formal proof and description of how this can be done.  As a result, $\partial \calE / \partial \sigma$ should generally be considered as a subgradient, and a choice of strict ordering selects an element in the subgradient.

{\bf Filtrations:} While general filtrations could be considered for optimization, we will focus on two different kinds of filtrations that are defined by either points or edges in a complex.  For simplicity, we now use simplicial complexes, where each cell is a simplex $(v_0,\dots,v_k)$, where each $(v_j)$ is a 0-cell (point).  We will use the subscript notation $\sigma_i$ to denote the $i$-skeleton of a simplex, which consists of the $i$-dimensional faces.  For instance $\sigma_0 = \{(v_j) \mid v_j \in (v_0,\dots, v_k) = \sigma\}$, and $\sigma_1$ consists of all $\binom{k}{2}$ pairs of 0-cells.

\begin{figure}[th]
    \vspace{-0.3cm}
    \centering
    \includegraphics[width=0.32\columnwidth]{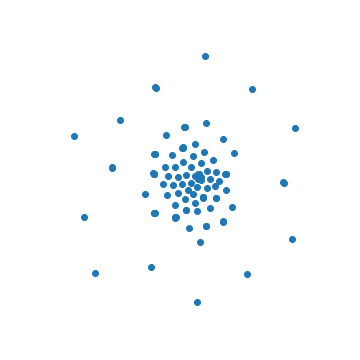}
    \includegraphics[width=0.32\columnwidth]{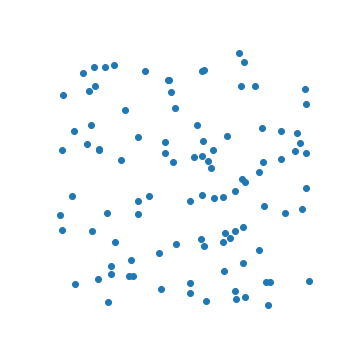}
    \includegraphics[width=0.32\columnwidth]{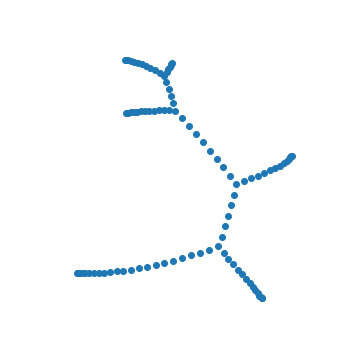}\\
    \includegraphics[width=0.32\columnwidth]{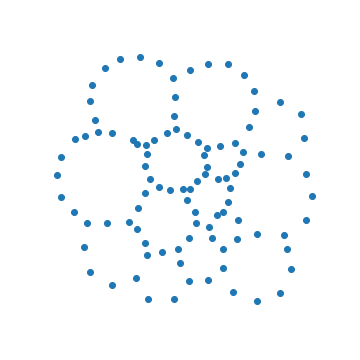}
    \includegraphics[width=0.32\columnwidth]{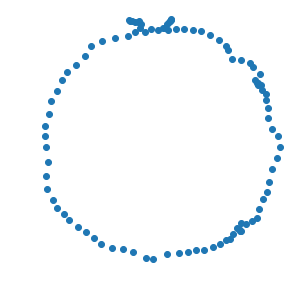}
    \includegraphics[width=0.32\columnwidth]{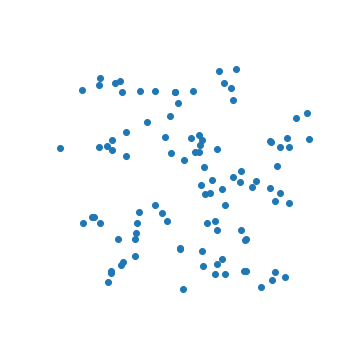}
    \caption{weak Alpha filtrations. Top center: points sampled uniformly from the unit square, Top left: Optimizing to increase $\mathcal{E}(2,0,2;\PD_0)$, Top right: Optimizing to decrease $\mathcal{E}(2,0,2;\PD_0)$.  Bottom Left: Optimizing to increase $\mathcal{E}(2,0,1;\PD_1)$, Bottom Right: optimizing to decrease $\mathcal{E}(2,0,1;\PD_1)$, Bottom center: optimizing to increase $\mathcal{E}(2,1,1;\PD_1)$, decrease $\mathcal{E}(2,0,2;\PD_0)$}
    \label{fig:alpha_optim}
\end{figure}
\begin{figure}[h]
     \vspace{-0.2cm}
     \centering
     \includegraphics[width=1.0\columnwidth]{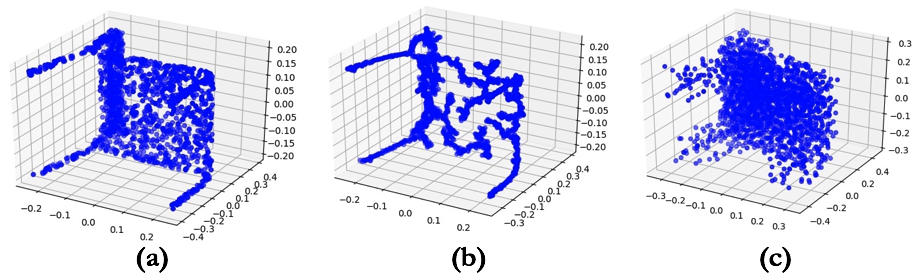}
      \vspace{-0.3cm}
     \caption{
     Rips filtrations. (a) points sampled from a chair (b) optimized to decrease $\mathcal{E}(1,0,2; \PD_0)$, (c) increase $\mathcal{E}(1,0,2; \PD_0)$.}
     \label{fig:ripsnoise}
     \vspace{-0.2cm}
 \end{figure}

First, we consider extensions of filtrations on 0-cells, also known as lower-star filtrations.  In particular for sublevel set filtrations, $f((v_0,\dots, v_k)) = \max_{i=0,\dots,k} f((v_i))$.  This construction is useful for building filtrations on images, where we take $\X$ to be a triangulation of a rectangle with 0-cells defined by the grid of pixels on the image, and $f((v_i))$ is the intensity of a color channel at the pixel $(v_i)$.

The second kind of filtration that we consider extends a filtration on the edges of a complex, also called a flag filtration.  For sublevel set filtrations, this has the form $f((v_0,\dots, v_k)) = \max_{i<j\in 0,\dots, k} f((v_i, v_j))$
One example of this is based on pairwise distances of points.  The Vietoris-Rips, or Rips, filtration, $\mathcal{R}_\alpha$, is the distance-based flag filtration on the clique complex, consisting of all $2^n$ possible simplices on the vertex set.  Even when limiting the space to simplices below a certain dimension, the Rips filtration can become too large to compute with efficiently.  A more tractable complex in low dimensional euclidean space uses the Delaunay triangulation of a point cloud as the underlying space.  We refer to the distance-based flag filtration on this space as the {\em weak Alpha filtration}.

 \begin{figure}[t]
   \centering 
   \includegraphics[width=0.99\columnwidth]{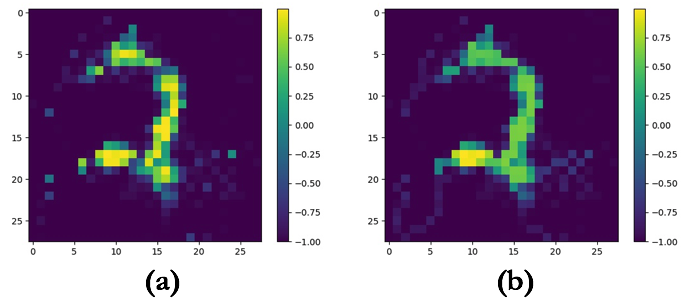}
   \vspace{-0.3cm}
  \caption{(a) A noisy image of the digit `2'. (b) after promoting a single local maximum using $\mathcal{E}(1,0,2; \PD_0).$}
  \label{fig:levelsetnoise}
\end{figure}
%
\begin{figure}[h]
\centering
\begin{minipage}{\columnwidth}
\centering
\begin{tabular}{|c||c|c|c|}
\hline
 Penalty & L1 &L2 &TV  \\
 \hline
 Def. &$\|\beta\|_1$ & $\|\beta\|_2$& $\|\nabla \beta\|_1$ \\
 \hline
 \hline
 Penalty&TV2&Top1 &Top2\\
 \hline
  Def.& $\|\nabla \beta\|_2$& $\mathcal{E}(1,0,2;\PD_0)$& $\mathcal{E}(1,0,4;\PD_0)$ \\
 \hline
\end{tabular}

\caption{Abbreviation for penalties used for regularization of least squares problems.}
\label{tab:penalties}
\end{minipage}
 \begin{minipage}{1.0\columnwidth}
     \centering
     \includegraphics[width=0.49\columnwidth]{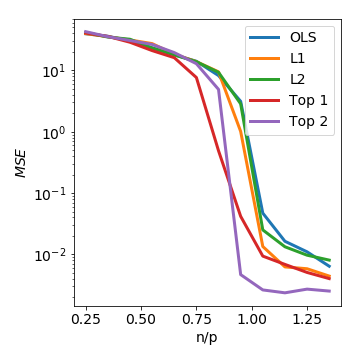}
     \includegraphics[width=0.49\columnwidth]{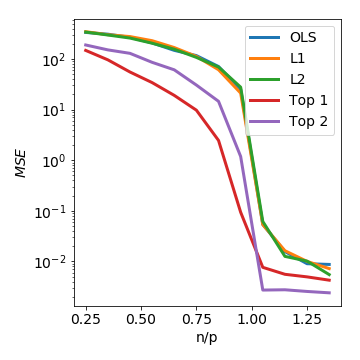}
     \caption{MSE of $\hat{\beta}$ obtained using several regularization schemes as size of training set increases.  Left: entries of $\beta_\ast$ are drawn i.i.d. from $\{-1,0,1\}$. Right: entries of $\beta_\ast$ are drawn i.i.d. from $\{1,2,3\}$.}
    \label{fig:regs_alpha}
 \end{minipage}
      \vspace{-0.2cm}

 \end{figure}

{\bf Computation:} We have implemented a PyTorch \citep{pytorch} extension that performs the described differentiation through persistence diagrams, supporting several standard algorithms written in C++. The actual method used to compute persistent homology is largely irrelevant for our purposes, as long as we are able to map points in the persistence diagram back to filtration values of individual cells.  While our implementation does not rely on external topology libraries, many existing packages could potentially be used or modified to provide the required information.  The original persistence algorithm~\citep{ZCComputingPH2005} as well as the cohomology algorithm~\citep{dualitiescohom} are based on putting the boundary matrices $\partial_k$ in a form which reveals the birth-death pairs. 
The worst case complexity is known to be equivalent to matrix multiplication in the number of simplices~\citep{ZZmatmultime2011}, although sparsity of $\partial_k$ typically renders this bound pessimistic.  There are many approaches to speeding up calculations in practice~\citep{otter2017roadmap}, and if only zero-dimensional homology is of interest, then the union-find algorithm~\citep{CLRS} typically performs faster.  The dependence of number of simplices on the number of points $n$ depends on the construction. For example, the Alpha complex may have $O(n^{d/2})$ simplices where $d$ is the ambient dimension \citep{alpha_survey}, whereas the Rips complex may have as many as $O(n^{k+1})$ simplices where $k$ is the maximal dimension homology we consider.  However, in practice, the resulting complexes are approximately linear in $n$ for small $d$ and $k$.
We note that many improvements to the persistence algorithm have been made with the goal of tackling larger spaces.  In contrast, we seek to compute persistence using different filtrations on the same small- to medium-sized space rapidly, which may find different optimizations beneficial, although these considerations are beyond the scope of this work.

\section{Applications}

\subsection{Topological Noise Reduction and Regularization}
\label{sec:noise}

We first demonstrate how functions of persistence diagrams can be effectively used for both optimization of the placement of points, and optimization of functions on a space.  We show how one can encourage the formation of lines, clusters, or holes in a set of points using geometric filtrations.  We then show how level set filtrations can be used effectively for regularization of parameters in a model by penalizing the number number of local maxima in the parameter topology.  While we see there is some benefit to regularization using an appropriate topological penalty, we do not claim superiority to other regularization schemes.  Instead we wish to draw attention to the flexibility of topological penalties in both the point cloud and image settings.
 
In Section \ref{sec:intro}, we reviewed several applications which use specific topological loss functions.  There are many possible losses which may be considered, and here we demonstrate some behaviors that can be promoted using persistence. 
In Figure \ref{fig:alpha_optim}, we see how a set of 100 random points in the unit square can be moved into different configurations by taking gradients of different functions of weak Alpha persistence diagrams.  In Figure \ref{fig:ripsnoise} we see how points that are sampled from a 3D chair can be moved around using similar functions of Rips persistence diagrams.  An analysis of the optimality of one choice over another in any given situation is beyond the scope of this work.  We primarily wish to draw attention to the wide variety of behaviors that can be encouraged by varying the choice of function.

 \begin{figure*}[t]
    \centering
    \includegraphics[width=0.24\textwidth]{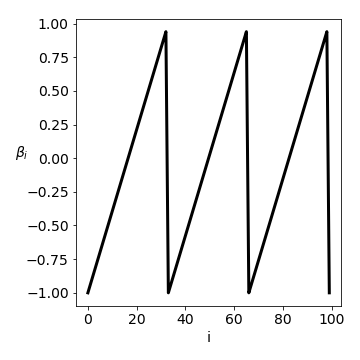}
    \hfill
    \includegraphics[width=0.24\textwidth]{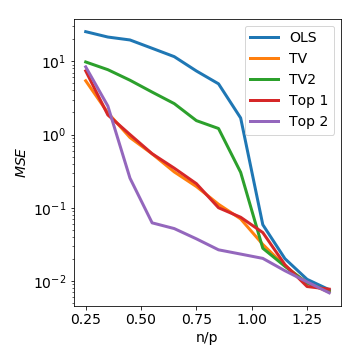}
    \hfill
    \includegraphics[width=0.24\textwidth]{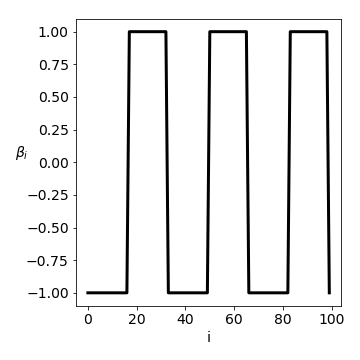}
    \hfill
    \includegraphics[width=0.24\textwidth]{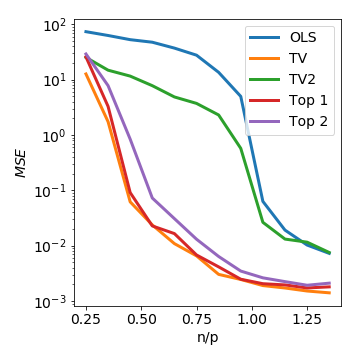}
    \caption{Left to right: Sawtooth $\beta_\ast$. MSE of linear prediction using $\hat{\beta}$ obtained from several regularization schemes as size of training set increases.  Boxcar $\beta_\ast$. MSE of linear prediction using $\hat{\beta}$ obtained from same regularization schemes as size of training set increases.}
    \label{fig:regs_levelset}
 \end{figure*}
  
\begin{figure}[h]
\begin{minipage}{1.0\columnwidth}
 \centering
 \includegraphics[width=1.0\columnwidth]{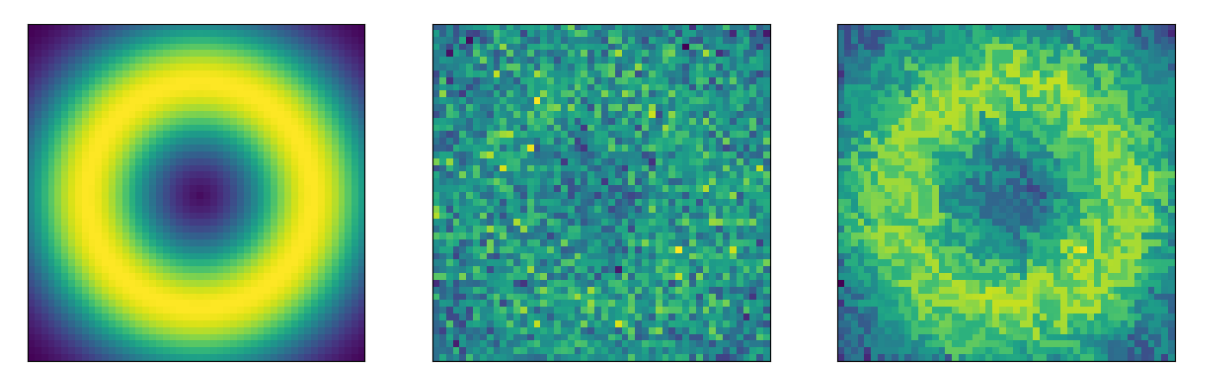}
 \caption{Left: Original image with pixel values in $[0,1]$.  Data is generated with $n/p = 0.5$, with i.i.d. Gaussian noise.  Center: ordinary least squares solution.  Right: least squares solution with topological penalty $\calE(1, 0, 2; \PD_0) + \calE(1, 0, 2; \PD_1)$. }
 \label{fig:noisy_circle}
      \vspace{-0.2cm}
\end{minipage}
\end{figure}
Direct optimization on the filtration is not limited to geometric complexes. In Figure \ref{fig:levelsetnoise}, we optimize functions on a space. As we will see in Section \ref{sec:top_loss}, limiting the number of local maxima in an image can improve the visual quality of generated digits.  In this example, we perform optimization directly on the superlevel sets of a noisy image to produce a single global maximum.

While these examples are illustrative, we wish to see how we can use topology directly in a machine learning model for the purposes of regularization, or encoding a prior on some topological structure.

\begin{figure*}[ht!]
     \centering
     \includegraphics[width=0.8\textwidth]{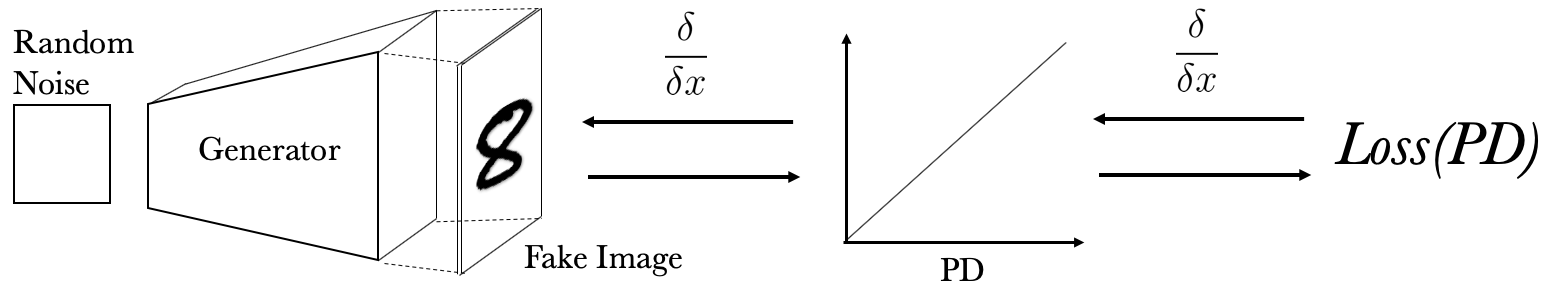}
     \includegraphics[width=0.8\textwidth]{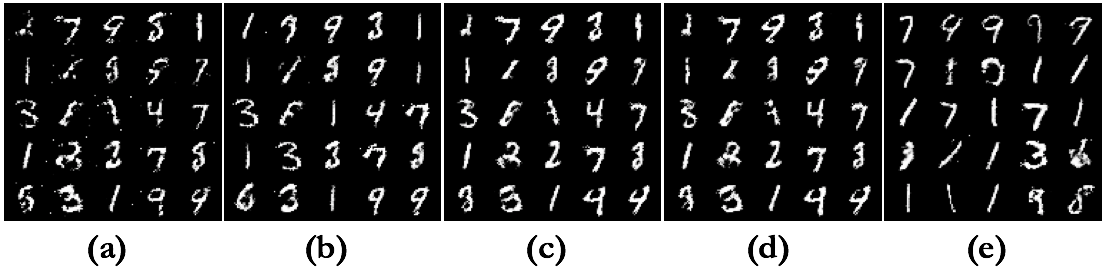}
     \caption{Top: Setup for computing topology loss and backpropogation. (a) Baseline-Generator. (b) Minimize the topology loss with respect to the latent space of Baseline-Generator. (c) Topology-Generator. (d) Train Baseline-Generator with topology-discriminator for 100 batch iterations. (e) Train Baseline-Generator in original GAN-setup for another 60,000 batch iterations.}
     \label{fig:costall}
 \end{figure*}
   \begin{figure}[h!]
     \centering
     \includegraphics[width=1\columnwidth]{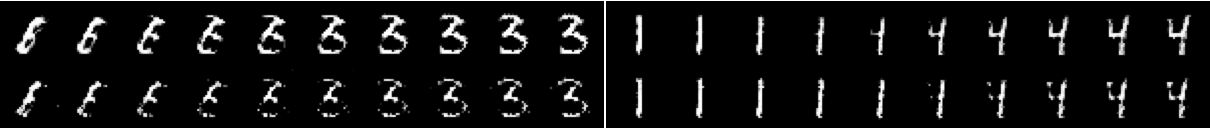}
     \caption{Linear interpolation in latent space. Top row: Topology-Generator. Bottom row: Baseline-Generator. }
     \label{fig:interpolation}
 \end{figure}
Regularization is used throughout machine learning to prevent over-fitting, or to solve ill-posed problems. In a typical problem, we observe data $\{X_i\}$ and responses $\{y_i\}$, and would like to fit a predictive model with parameters $\hat{\beta}$ that will allow us to make a prediction $\hat{y_i} = f(\hat{\beta}; X_i)$ for each observation.  The quality of the model is assessed by a loss function $\ell$, such as the mean squared error.  However, many models are prone to over-fitting or are ill-posed if there are more unknown parameters than observations, and adding a regularization term $P(\beta)$ can be beneficial. The estimated value of $\hat{\beta}$ for the model becomes
\begin{equation}
    \hat{\beta} = \operatorname*{argmin}_{\beta} \sum_{i=1}^n \ell\big(y_i, f(\beta; X_i)\big) + \lambda P(\beta)
    \label{eqn:regularization}
\end{equation}
where $\lambda$ is a free tuning parameter.

Well-known examples of regularization include $L_1$ regularization $P(\beta) = \|\beta\|_1$ (Lasso) \citep{lasso}, which promotes sparsity, or $L_2$ regularization $P(\beta) = \|\beta\|_2$ (Ridge regression) \citep{ridge} which tends to keep parameters from growing excessively large.  Both of these types of regularization can be viewed as making the topological statement that parameter weights should ``cluster'' around zero, and a similar topological penalty might simply encourage the set of all weights to form clusters by penalizing the sum of lengths of $\PD_0$ from a Rips or weak Alpha filtration on the weights.

Another class of well-known regularization schemes  make an assumption about the topology of the set of parameters themselves, and penalize properties of the weights as a function on that space.  Examples include penalties on a norm of a finite-difference derivative, such as total variation regularization $P(\beta) = \|\nabla \beta\|_1$ \citep{TVReg}, or penalties on the ordering of weights as seen in isotonic regression and its variants \citep{nearlyisotonic}.  From the topological point of view, these regularization schemes encourage $\beta$ to have fewer local maxima and minima, which might be accomplished by penalizing the sum of lengths of $\PD_0$ from a level set filtration.

In Figures \ref{fig:regs_alpha} and \ref{fig:regs_levelset}, we compare different regularization schemes for several different linear regression problems.  Examples are generated according to $y_i = X_i\beta_\ast + \epsilon_i$, with $X_i\sim N(0,I)$, and $\epsilon_i\sim N(0,0.05)$.  $\beta_\ast$ is a feature vector with $p=100$ features, and an estimate $\hat{\beta}$ is made from $n$ samples by solving Equation \ref{eqn:regularization} with the mean-squared error loss $\ell\big(y_i, f(\beta; X_i)\big) = (y_i - X_i \beta)^2$ using different penalties, and $\lambda$ is chosen from a logarithmically spaced grid on $[10^{-4},10^1]$ via cross-validation for each penalty.  We track the mean-squared prediction error for the estimate $\hat{\beta}$ as the number of samples is increased.  We also compare to the ordinary least-squares solution, using the smallest 2-norm solution if the problems is under-determined $(n < p)$.

In Figure \ref{fig:regs_alpha}, $\beta_\ast$ are chosen uniformly at random from three different values.  On the left, those values are $\{-1,0,1\}$, and on the right, $\{1, 2, 3\}$.  We consider $L_1$ and $L_2$ penalties, as well as two topological penalties using a weak-alpha filtration.  The first is $\mathcal{E}(1,0,2; \PD_0)$, and the second is $\mathcal{E}(1,0,4; \PD_0)$.  Both topological penalties are non-negative, and the first penalty is non-zero if $\beta$ takes more than a single value, and the second penalty is non-zero if $\beta$ takes more than three distinct values,  explicitly encoding that we expect three clusters.  In the case where $\beta_\ast$ takes values in $\{-1,0,1\}$, the $L_1$ and $L_2$ penalties slightly outperform ordinary least squares, because while $\beta_\ast$ is not truly sparse, some shrinkage seems beneficial.  In the case where $\beta_\ast$ takes values in $\{1,2,3\}$, $L_1$ and $L_2$ clearly bias the estimate in an ineffective way and fail to outperform ordinary least squares.  In contrast, the two topological penalties clearly do better in both cases.

In Figure \ref{fig:regs_levelset}, the features in $\beta_\ast$ are chosen to have three local maxima when the features are given the line topology. On the left, $\beta_\ast$ consists of three piecewise-linear sawteeth, and on the right, $\beta_\ast$ consists of three piecewise-constant boxcars.  The total variation penalty $P(\beta) = \sum_{i=1}^p |\beta_{i+1} - \beta_i|$ and a smooth variant $P(\beta) = (\sum_{i=1}^p |\beta_{i+1} - \beta_i|^2)^{1/2}$ are considered, as well as two topological penalties.  The parameters of the topological penalties are identical to the previous example, but the penalties are now imposed on superlevel set diagrams of $\beta$ in order to penalize the number of local maxima in $\beta$ instead of the number of distinct values.  In the boxcar problem, total variation regularization does very well, as it encourages piece-wise linear functions, and the two topological penalties perform similarly.  In the sawtooth problem, total variation does not do as well because $\beta_\ast$ is no longer piece-wise constant, and interestingly the first topological penalty is similarly not as effective, while the second topological penalty performs well in both examples.

Finally, Figure \ref{fig:noisy_circle} shows a linear regression problem on a 2D image.  The topological penalty incorporated information from $\PD_1$ as well as $\PD_0$ to promote a single maximum and a single hole.  For visual comparison, we also show the resulting ordinary least squares image.

These examples demonstrate how topological information can be incorporated effectively to add regularization or incorporate prior knowledge into problems. Furthermore, they demonstrate how topological information can be directly encoded, such as penalties on the number of clusters or number of maxima of a function, in a natural way that is difficult to accomplish with more traditional schemes.  

\subsection{Incorporating Topological Priors in Generative Models}
\label{sec:top_loss}

 \begin{figure*}[t]
\centering
\begin{tabular}{ |p{5.1cm}||p{1.5cm}|p{1.8cm}|p{1.9cm}|p{1.8cm}|p{1.5cm}|  }
 \hline
 \multicolumn{6}{|c|}{Generator Evaluation} \\
 \hline
 Model & MMD-L2 & COV-L2 & MMD-Wass & COV-Wass & Inception \\
 \hline
  Baseline-Generator (Images) & 28.0$\pm$0.1 & 0.05$\pm$0.006 & 1.56$\pm$0.08 & 0.11$\pm$0.01 & 4.6$\pm$0.1 \\
 Topology-Generator (Images) & \textbf{27.5}$\pm$0.1 & \textbf{0.06}$\pm$0.002 & \textbf{1.52}$\pm$0.07 & \textbf{0.12}$\pm$0.01 & \textbf{5.1}$\pm$0.1 \\
 \hline
 Baseline-Generator (3D Voxels) & \textbf{33.7}$\pm$0.8 & 0.10$\pm$0.01 & 4.3$\pm$1.4 & 0.65$\pm$0.03 & N/A \\
 Topology-Generator (3D Voxels) & 34.1$\pm$0.8 & \textbf{0.11}$\pm$0.02 & \textbf{2.4}$\pm$0.8 & 0.65$\pm$0.03 & N/A \\
 \hline
\end{tabular}
\caption{Metrics for generator evaluation.}
\label{tbl:gentable}
 \end{figure*}
\begin{figure}[h]
     \centering
     \includegraphics[width=0.99\columnwidth]{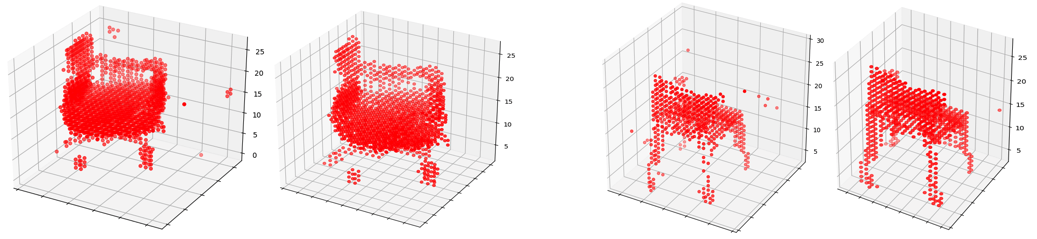}
     \caption{For each pair: (Left) before training with topology loss, (Right) after training with topology loss for 20 batch iterations.}
     \label{fig:cost3d}
          \vspace{-0.4cm}
 \end{figure} 

 We now use the same topological priors to improve the quality of a deep generative neural network. We start with a Baseline-Generator, pre-trained in a GAN-setup on MNIST, and by training it for a few iterations with a topological loss, we arrive at an improved Topology-Generator. We provide comparisons with other methods applied to the Baseline-Generator.

A GAN as in~\citep{GAN} is trained on MNIST for 32,000 batch iterations with a batch size of 64 (this batch size is used throughout this section). The resulting generator (Baseline-Generator) produces reasonable output but with topological noise, see Figure \ref{fig:costall}(a). The prior used to improve the Baseline-Generator is identical with that of Figure \ref{fig:levelsetnoise}: images should have 1 component in a superlevel set filtration. The loss function (topology loss) is $\mathcal{E}(1,0,2; \PD_0)$ .
The setup (Figure \ref{fig:costall}) is used to backpropagate to the latent space of Baseline-Generator, with the generator weights fixed, to minimize the topology loss using SGD; seen Figure \ref{fig:costall}(b) for results. ALternatively, using the same setup, the Baseline-Generator's weights are updated to minimize the topology loss; we train for 50 batch iterations to arrive at a new generator (Topology-Generator).  The output can be seen in Figure \ref{fig:costall}(c). 

For further qualitative comparisons, we train the Baseline-Generator for 100 batch iterations with a discriminator between features of the 0-dim persistence on MNIST images and the generator's output. The features were sums of lengths of the $k$ longest $\PD_0$ features for several choices of $k$, expressed as $\mathcal{E}(1, 0, 2; \PD_0)-\mathcal{E}(1, 0, 3; \PD_0)$, $\mathcal{E}(1, 0, 2; \PD_0)-\mathcal{E}(1, 0, 4; \PD_0)$, $\mathcal{E}(1, 0, 2; \PD_0)-\mathcal{E}(1, 0, 5; \PD_0)$, and $\mathcal{E}(1, 0, 2; \PD_0)-\mathcal{E}(1, 0, 11; \PD_0)$, with the results shown in Figure \ref{fig:costall} (d). The output of a generator arrived at by training the Baseline-Generator in the original GAN-setup for another 60,000 batch iterations is shown in Figure \ref{fig:costall} (e).
Evidently, the topology loss allows the generator to learn in only 50 batch iterations to produce images with a single connected component and the difference is visually significant. These results are similar to using a $\PD_0$-aware discriminator, suggesting that our priors were valid. Updating only the latent space produces cleaner images but they still contain some topological noise. For a closer study, consider the linear interpolation in the latent space of the Baseline-Generator and Topology-Generator in Figure \ref{fig:interpolation}. The two different cases behave very differently with respect to the topology. The Baseline-Generator interpolates by letting a disconnected components appear and grow.  The Topology-Generator tries to interpolate by deforming the number without creating disconnected components. This might be most obvious in the interpolation from ``1'' to ``4'' (Figure \ref{fig:interpolation}, right hand side) where the appended structure of the ``4'' appears as a disconnected component in the baseline but grows out continuously from the ``1'' in the topology-aware case.

 \begin{figure*}[t]
     \centering
     \vspace{-0.2cm}
     \includegraphics[width=0.3\textwidth]{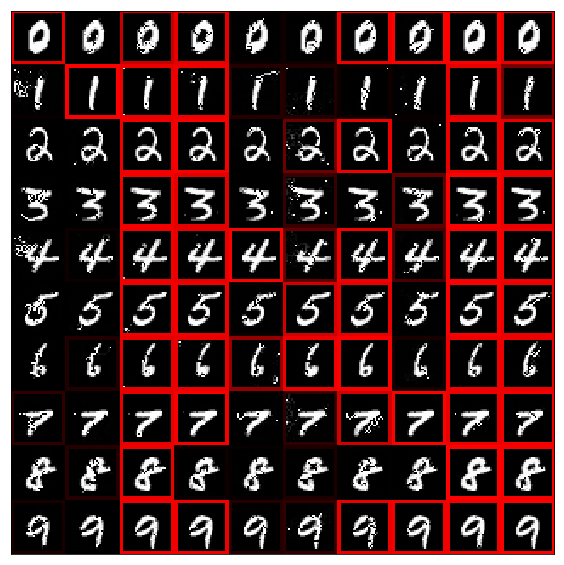}
     \hfill
     \includegraphics[width=0.3\textwidth]{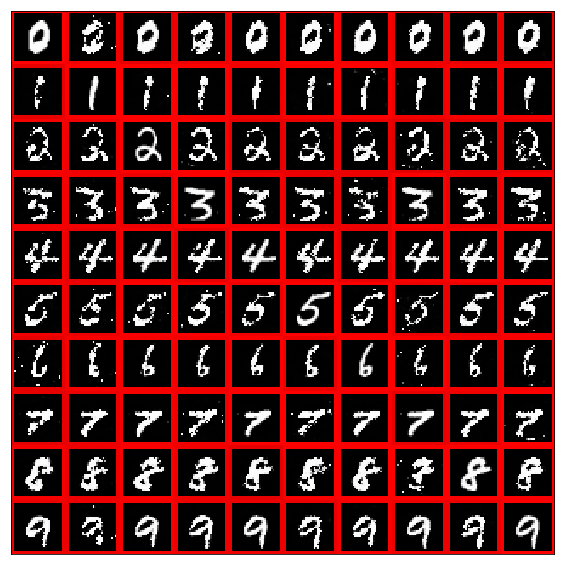}
     \hfill
     \includegraphics[width=0.3\textwidth]{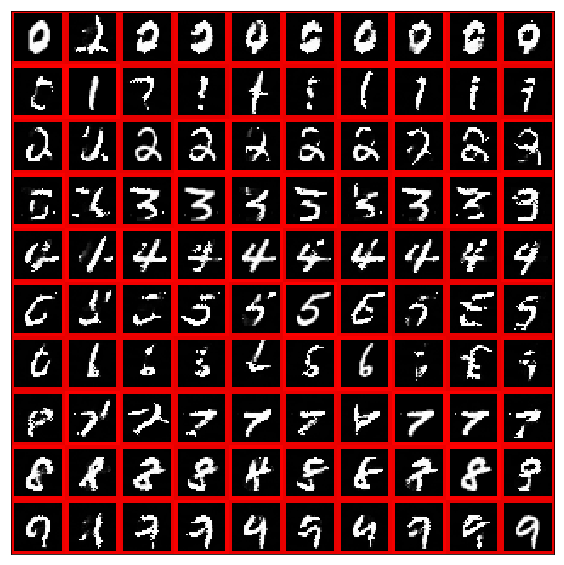}
     \caption{Topological adversarial attack on TopModel, MLPModel and CNNModel. The $(i,j)$-th cell 
     represents an attack on an image with label $i$ to be classified as label $j$. Red outlines are successful attacks.}
     \label{fig:attacks}
 \end{figure*}
 \begin{figure}[h!]
   \centering 
   \includegraphics[width=0.35\columnwidth]{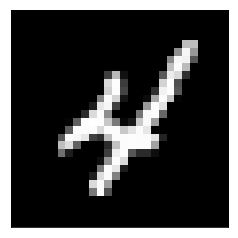}
   \hspace{0.5cm}
    \includegraphics[width=0.35\columnwidth]{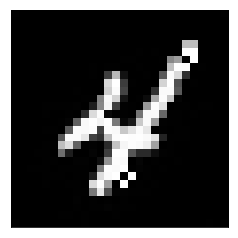}
   \vspace{-0.1cm}
  \caption{Example of Topological adversarial attack. (Left) original image, (Right) image  optimized to be classified as an 8, which introduced two 1 pixel holes.}
 \vspace{-0.5cm}
  \label{fig:attack4to8}
 \end{figure}

We also quantitatively compare the Baseline-Generator and Topology-Generator to further investigate if any improvements have been made. We use the Minimal Matching Distance (MMD) and Coverage metric as advocated by~\citep{Panos} as well as the \textit{Inception score}~\citep{Inception} (a convolutional neural network with 99\% test accuracy on MNIST was used instead of the Inception model).  MMD-Wass and COV-Wass use the same procedure as MMD-L2 and COV-L2 but instead of the L2 distance between images, the 1-Wasserstein distance between the 0-dim persistence diagrams of the images was used (see Section \ref{sec:persistence}). As seen in Table \ref{tbl:gentable}, the Topology-Generator shows improvements for all of these metrics. The results are the average of 5 computations of each metric, with test set sizes of 1,000 for L2 and Inception, and test sets sizes of 100 for Wasserstein distance.

 We extend this superlevel set filtration to 3D data in the form of voxel grids. As before, a baseline generator is obtained by training a GAN to generate voxel shapes (chairs only) as in~\citep{3dgan} and its output after 1,000 epochs (or 333,000 batch iterations) can be seen in Figure \ref{fig:cost3d} as the left hand members in each of the two pairs. The result of training with the topology loss (same as for images) for 20 batch iterations can be seen in Figure \ref{fig:cost3d} as the right hand members in each of the two pairs. We compare some metrics in Table \ref{tbl:gentable}; we show the average of 5 computations of each metric, with test set sizes of 100. Note that every voxel chair in the ground truth dataset has identical $\PD_0$, since each chair consists of a connected component of voxels of value 1, among voxels of value 0.

\subsection{Topological Adversarial Attacks}

Our topological layer may also be placed at the beginning of a deep network to generate features directly on the data. We can use the fact that our input layer is differentiable to perform adversarial attacks, by backpropagating from the predictions back to the input image. To the best of our knowledge, these are the first adversarial attacks conducted using persistence features.
Since standard super-level set persistence is insufficient to classify MNIST digits, we include orientation information by computing the persistence homology during 8 directional sweeps. This is achieved by using the product of the image with fixed functions such as $x$, $y$, $\frac{x+y}{2}, \dots$ etc., where $x$ and $y$ are the image coordinates, as the filtration value, for each of 8 different directions, $\mathcal{E}(p, q, 1; \PD_k)$ for $p$ and $q$ ranging between 0 and 4 resulting in 400 features for training the classification model.
The model trained to classify the digits based on these topological features achieved 80-85 \% accuracy. Next we performed gradient attack~\citep{goodfellow2014explaining} to change the classification of the digit to another target class. We observe that it is harder to train adversarial images compared to CNNs and MLPs. The results are shown in Figure \ref{fig:attacks}. A red outline indicates that the attack was successful. When the attack was conducted on 1,000 images, to retarget to a random class, it had 100\% success rate on MLP and CNN models and 25.2\% success rate on the TopModel.  
When the adversarial attacks succeed, the results may offer insight on how the model classifies each digit. For example in Figure \ref{fig:attack4to8}, the left image is the original image of the digit 4, the right was trained to be classified as an 8; note that two small holes at the top and bottom were sufficient to misclassify the digit. Several examples of the topological attacks provide similar intuition. Attacks on MLP and CNN are qualitatively different, but further work is needed to gauge the extent and utility of such distinctions.

\section{Discussion}
We present three novel applications using a differentiable topology layer which can be used to promote topological structure in Euclidean data, images, the weights of machine learning models, and to compare adversarial attacks. 
%
%
%
This only scratches the surface of the possible directions leveraging the differentiable properties of persistence. Without doubt such work will tackle problems beyond those we have presented here, including encouraging topological structure in intermediate activations of deep neural networks or using the layer in the middle of deep networks to extract persistence features where they may be more useful. However, many of the applications we have presented here also deserve further focus. For example, topological regularization, including the penalties we have presented, may have interesting theoretical properties, or closed form solutions.  Furthermore, training autoencoders with distances between persistence features may produce stronger results than the functions considered here. Finally, it might prove useful to use topological features to train deep networks that are more robust to adversarial attacks -- however, as we show this will require additional work. Topology, in contrast to local geometry, is generally underexploited in machine learning, but changing this could benefit the discipline.

\noindent \textbf{Acknowledgements:} RBG and GC were supported by Altor Equity Partners AB through Unbox AI (\url{www.unboxai.org}).  BN was supported by the US Department of Energy (DE-AC02-76SF00515) while at the SLAC National Accelerator Laboratory.  PS was supported by SSHRC Canada (NFRFE-2018-00431) and the Alan Turing Institute - Defense and Security Programme (D015).  We are grateful for Panos Achlioptas’ insights on the evaluation of our generative models.


\printbibliography

\clearpage
\appendix
\section{Appendix}

\subsection{Topological Definitions}
Here we include the precise definitions of the required notions -- this should not be considered as an exhaustive introduction, but rather our goal is to give precise definitions to the concepts used in the paper. 

A cell complex can be defined inductively by dimension.  Let $\X_0$ be the 0-skeleton of the complex consisting of points, which can be thought of as 0-dimensional balls (cells).  The $k$-skeleton $\X_k$ is obtained by attaching some number of $k$-dimensional cells by their boundaries to the $(k-1)$-skeleton $\X_{k-1}$ via continuous maps.  This process can be repeated indefinitely to form the cell complex $\X$, but in practice we terminate at some finite dimension.  While cell complexes can be used fairly flexibly in encoding topological spaces, simplicial complexes are typically preferred in computational settings for their combinatorial description.

\begin{definition}
A $k$-dimensional simplex $(v_0,\dots,v_k)$ is the convex combination of $k+1$ vertices, $\{(v_0),\dots,(v_k)\}$.
\end{definition}
We only consider the situation where the vertices correspond to point in a sufficiently nice ambient space, usually $\mathbb{R}^d$ so that a geometric realization of simplices exists as a subset of the ambient space. Simplices are used to represent a space, but must satisfy some additional constraints so that they form a \emph{simplicial complex}. 
\begin{definition}
A simplicial complex $\X$ is a collection of simplices such that 
\begin{enumerate}
    \item For every simplex $\sigma$ in $\X$, every face $\tau\subseteq \sigma$ is also in $\X$.
    \item For any two simplices $\sigma_1$ and $\sigma_2$, $\tau = \sigma_1\cap \sigma_2$ is a face of both $\sigma_1$ and $\sigma_2$. 
\end{enumerate}
\end{definition}
These can be thought of as higher dimensional analogs of graphs or triangular meshes -- conversely, a graph is a 1-dimensional simplicial complex.

To define homology, we first construct a \emph{chain group}. In our setting, the $k$-th chain group $C_k(\X)$ is the freely generated group generated by $k$-dimensional simplices. Importantly, there exists a boundary homomorphism  
$$\partial_k:C_{k}(\X)\rightarrow C_{k-1}(\X) $$
such that $\partial\circ \partial = 0 $. Explicitly,
$$\partial_k (v_0,\dots,v_k) = \sum_{i=0}^k (-1)^i (v_0,\dots,\hat{v}_i,\dots, v_k)$$
where $\hat{v}_i$ indicates that the $i$-th vertex has been removed.  Again, we take $\partial_0 = 0$.
Homology  can then be defined as 
$$H_k(\X) = \frac{\mathrm{ker}\; \partial_k}{\mathrm{im}\; \partial_{k+1}} $$
We consider homology computed over fields, in which case the homology groups are vector spaces.  Rather than only consider one simplicial complex $\X$, we can consider an increasing sequence of simplicial complexes, called a \emph{filtration} $\X_0\subseteq \X_1 \subseteq \ldots \subseteq \X_N$. The requirement is that each $\X_i$ is itself a simplicial complex. We consider the filtration function $f:\X\rightarrow \mathbb{R}$ such that each sub/super-level set $f^{-1}(-\infty,\alpha]$, resp. $f^{-1}[\alpha,\infty)$ form a filtration in parameterized by $\alpha$. Under appropriate finiteness conditions, which are always satisfied for finite simplicial complexes~\citep{books/daglib/0025666}.
\begin{theorem}\citep{ZCComputingPH2005}
If homology is computed over a field, then the homology of a filtration admits an \emph{interval decomposition}. The interval decomposition is direct sum of rank 1 elements which exists over an interval $[b,d)$.
\end{theorem}
In our setting, whether the intervals are open or closed is not important so we supress this aspect of the notation. This collection of intervals is the \emph{persistence diagram}.
\begin{definition}
A persistence diagram is a collection of points  $(b_i,d_i)$, possibly with repetition, along with the diagonal $\Delta$, i.e. all points such that $b=d$.
\end{definition}
The diagonal plays an important part in the definition of distances between diagrams. The main distance we consider is the $p$-Wasserstein distance between two diagrams 
$$W_p(\PD_k(f),\PD_k(g))  = \inf\limits_{\varphi} \left(\sum\limits_{p\in \PD_k(f)} || p-\varphi(p)||^p\right)^{1/p} $$
where $\varphi$ is a bijection. 
This can be viewed as an optimal transport problem, points must either be matched to the other diagram or the diagonal. This can be formulated as a classical linear program if we add to each diagram the projection of the other diagram to the diagonal
$$\mathrm{proj}(d,b) = \left(\frac{d+b}{2},\frac{d+b}{2}\right) $$
In other words, we can consider the optimal transport problem between $\PD(f) \cup \mathrm{proj}(\PD(g))$
and $\PD(g) \cup \mathrm{proj}(\PD(f))$ where the distance between any two points on the diagonal is 0.

One of the core techniques in this paper is to compute the gradient of an energy function with respect to some parameters. This is possible through the definition of an inverse map
$$\pi_f(k) :   \{b_i,d_i\}_{i\in\mathcal{I}_k} \rightarrow (\sigma,\tau)$$
This map is unique and well defined in the case of when a filtration is a strict order, i.e. the simplices are totally ordered in the filtration. This is never the case in our scenarios, where the simplices only form a total order. However this can be resolved by extending the total order to a strict order either deterministically or randomly, see \citep{skraba2017randomly} for a formal proof and description of how this can be done.  

As is the case in our setting, the filtration is not defined directly on the simplices but rather derived from either functions on the vertices or depend on some property of the vertices, e.g. the coordinates of the points that the vertices correspond to. In this case, we require an additional inverse map from simplices to a collection of vertices. This is described in the main paper for two scenarios which we use in the applications. Once both inverse maps are defined, the gradient can be defined in the standard way using the chain rule.  

\begin{figure}
     \centering
     \includegraphics[width=0.99\columnwidth]{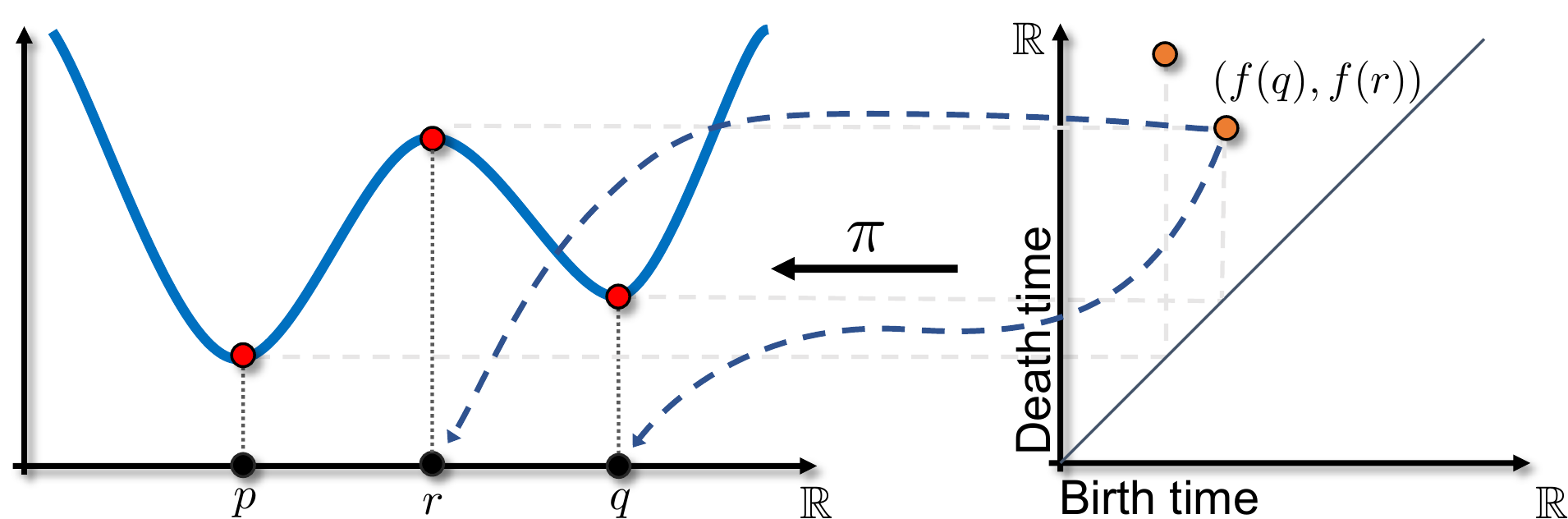}
     \caption{A one dimensional example of a persistence diagram and the inverse map $\pi$. The function on the left has critical points at points $p$, $r$ and $q$. The local minima create components in the sub-level sets and so represent birth times ($x$-axis), while the maxima kills one of the components (the younger one) and so is a death time ($y$-axis). The inverse map for a point in the diagram returns the corresponding critical points/simplicies. }
     \label{fig:overview}
 \end{figure}
\subsection{Comparison}

\begin{figure}[h]
     \centering
     \includegraphics[width=0.99\columnwidth]{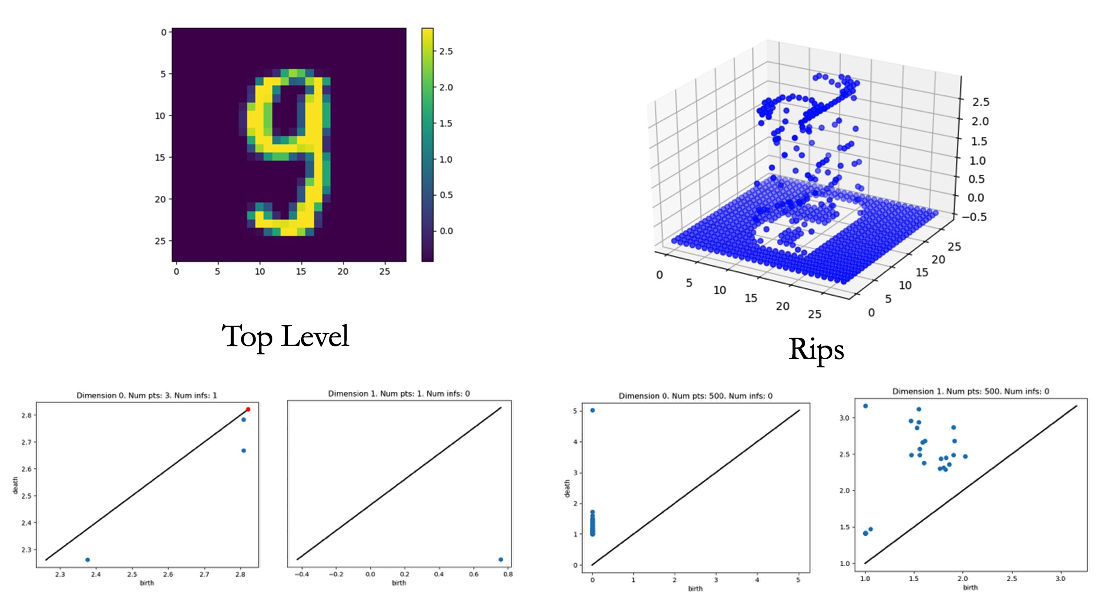}
     \caption{Compare top level set with Rips filtration}
     \label{fig:toprips}
 \end{figure}

In this paper we use both super-level set filtrations as well as Vietoris-Rips and weak-Alpha filtrations. Figure \ref{fig:toprips} provides an illustrative comparison of super-level set filtration (on the left) and a Rips filtration (on the right). The digit "9" viewed in terms of the super-level set of its pixel values has one clear connected component and one clear ring, which can be read from the corresponding persistence diagrams. However, the super-level set filtration does not take into consideration distance along the grid and a small (in the 2D sense) hole can be very large from the superlevel set perspective. In the Rips case we lay out all the pixel values in a 3D space and use the euclidean distance. Here there are many more connected components at small to medium filtration values, and the same is true for the number of rings. The Rips does not merely pick up the obvious ring in the "9" but also rings that are formed with respect to the vertical axis.

\subsection{Explore latent space}

\begin{figure}[h]
     \centering
     \includegraphics[width=0.99\columnwidth]{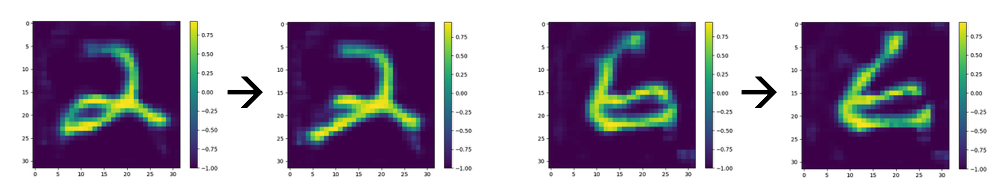}
     \caption{Infogan query}
     \label{fig:infogan}
 \end{figure}

In this paper we backpropagate to the latent space of generative models (query the latent space) in order to improve on them. However, the ability to topologically query the latent space may not only be used to give us more topologically desirable output but also to explore the nature of the latent space. Consider Figure \ref{fig:infogan} that consists of images produced by a trained InfoGAN-generator~\citep{Infogan}. The type of digit closest to a "2" with a loop on the bottom is arguably a digit "2" without such a loop. However, when we backpropagate to remove the ring we get something that does not look like any MNIST digit. Similarly, if we remove the ring in the digit "6" we get something that does not look very much like any MNIST digit. This gives some indications that the latent space learned by the InfoGAN is relatively topologically flexible.

 \subsection{Topological Mapping}
\begin{figure}[h]
     \centering
     \includegraphics[width=0.99\columnwidth]{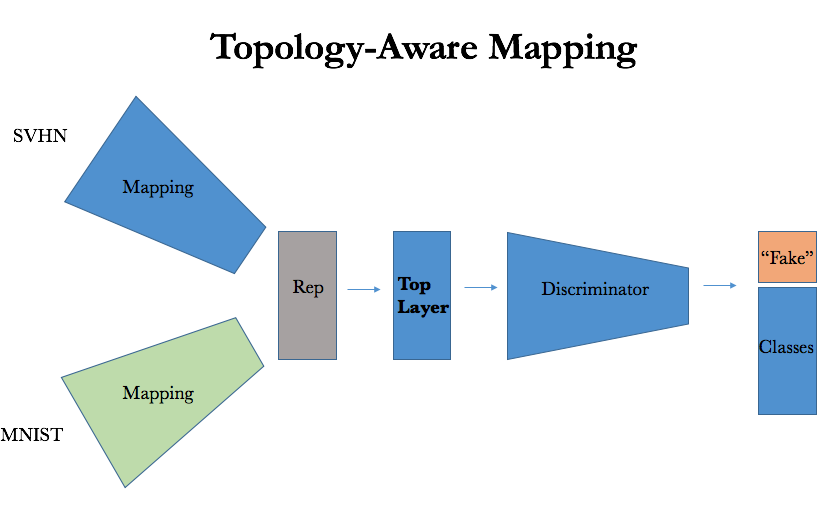}
     \caption{Topological mapping}
     \label{fig:mappingsetup}
 \end{figure}

\begin{figure}[h]
     \centering
     \includegraphics[width=0.99\columnwidth]{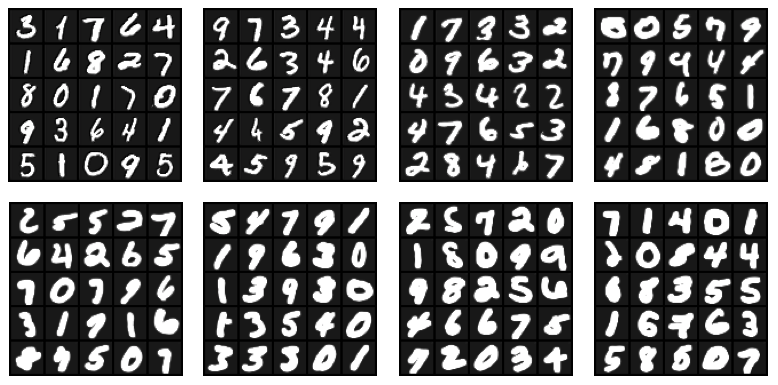}
     \caption{Intermediate Representation in topological mapping}
     \label{fig:mapping}
 \end{figure}

In this paper we train and experiment with networks that use persistence as input. However, with a differentiable layer we can put the persistence computations anywhere in the network. Sometimes it is useful to put certain layers in the middle of a deep network. For example in domain transfer we often want to create a mapping where features will be more useful. The setup might look like in the Figure \ref{fig:mappingsetup}.  
We do a simpler setup where we train a network to use superlevel persistence features to classify the naive $PD_1$-features of MNIST which we define for "1", "2", "3", "4", "5", "7" as zero, for "0", "6", "9" as one, and for "8" as two. We are able to get a 84\% accuracy. However, by putting two convolutional layers (the mapping) which maintain the dimensionality of the input before our Topology Layer, we get the results shown in Figure \ref{fig:mapping} where we ultimate improve our accuracy to 86\%.
In this case, it seems as if the mapping thickens the digits in the images in order to remove topological noise. 

\subsection{Topological Adversarial Attacks}
\subsubsection{Features for classification model}
In order to train a classification model using only topological features we must include orientation and directional information into the persistence homology features. This can be achieved by using custom filtration constructed in the following way. First we define 8 directional functions 
$$g_\theta(x,y) = \cos(\theta) x + \sin(\theta) y \;\;\;\;\;\theta = 0,\pi/4,\pi/2,\dots, 7\pi/4$$

These functions are shifted and scaled so that in the domain of the image, they range from 0 to 1. If $I(x,y)$ is the input image, then the filtrations are given by
$$ f_{\theta_i}(x,y) = I(x,y)g_{\theta_i}(x,y)$$
Persistence diagrams of dimensions 0 and 1 are computed for each filtration. We then compute 25 features on each persistence diagram given by $\mathcal{E}(p, q, 0; \PD)$, for $p$ and $q$ ranging between 0 and 4, thus totalling 400 features.

\subsection{Regularization Sparsity Visualization}

\begin{figure}[h]
    \centering
    \includegraphics[width=0.99\columnwidth]{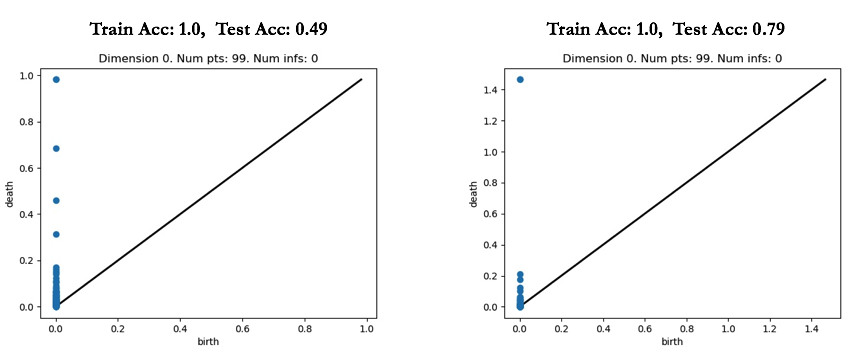}
    \caption{Rips and generalization}
    \label{fig:regcomp}
\end{figure}

In Figure \ref{fig:regcomp} we see the Rips $\PD_0$ diagram for the weights of a logistic regression model without (left) and with (right) L1 regularization. As the test accuracy improves, we see that the persistence diagram changes as well. 

\end{document}